# A Recursive Bateson-Inspired Model for the Generation of Semantic Formal Concepts from Spatial Sensory Data

Jaime de Miguel Rodríguez[*1], Fernando Sancho Caparrini[1a]

[1]Department of Computer Science and Artificial Intelligence, University of Seville,
Avda. Reina Mercedes, Seville, Spain

**Abstract.** Neural-symbolic approaches to machine learning incorporate the advantages from both connectionist and symbolic methods. Typically, these models employ a first module based on a neural architecture to extract features from complex data. Then, these features are processed as symbols by a symbolic engine that provides reasoning, concept structures, composability, better generalization and out-of-distribution learning among other possibilities. However, neural approaches to the grounding of symbols in sensory data, albeit powerful, still require heavy training and/or tedious labeling for the most part. This paper presents a new symbolic-only method for the generation of hierarchical concept structures from complex spatial sensory data. The approach is based on Bateson's notion of difference as the key to the genesis of an idea or a concept. Following his suggestion, the model extracts atomic features from raw data by computing elemental sequential comparisons in a stream of multivariate numerical values. Higher-level constructs are built from these features by subjecting them to further comparisons in a recursive process. At any stage in the recursion, a concept structure may be obtained from these constructs and features by means of Formal Concept Analysis. Results show that the model is able to produce fairly rich yet human-readable conceptual representations without training. Additionally, the concept structures obtained through the model (i) present high composability, which potentially enables the generation of 'unseen' concepts, (ii) allow formal reasoning, and (iii) have inherent abilities for generalization and out-of-distribution learning. Consequently, this method may offer an interesting angle to current neural-symbolic research. Future work is required to develop a training methodology so that the model can be tested against a larger dataset.

## 1. Introduction

The motivation behind this work is to explore new methods for the generation of rich and composable representations from spatial sensor data. In humans, these representations are very sophisticated in terms of their capacity to both differentiate and assimilate the realities they perceive. Additionally, they are extremely flexible, composable and feature logic-like operations. Therefore, the bridge between raw sensor data and composable internal models of the world, is key to artificial intelligence (AI). Symbolic or logic-based AI systems are said to emulate some of the important traits of human thought, such as compositionality, causality or the ability to reason (Newell, 1980; Simon, 1995). However, they face very significant challenges when dealing with complex raw data, such as, the symbol grounding problem (SGP) formulated by Harnad (Harnad, 1990), and a problem of computational complexity. Conversely, the connectionist approach seeded by some of the early cyberneticians (Wiener, 1948a) and the work of Pitts and McCulloch (Pitts & McCulloch, 1947), has proven exceptionally capable in dealing with raw data through neural network paradigms like deep learning (Lecun et al., 2015). Despite the success, connectionist models have some limitations when it comes to reasoning, flexibility and compositionality (which are more naturally handled by symbolic models) and out-of-distribution learning. Important authors in the field have made explicit statements of these aspects (Bengio, 2017; Lake et al., 2015; Xia et al., 2021), and very important efforts are being made to overcome them. Currently, a strong push in this direction is being led by neural-symbolic models (Garcez et al., 2015), and the literature around them has grown significantly in the last few years (Hitzler et al., 2022). These models incorporate methods from the two areas of AI discussed above, and that have been traditionally somewhat segregated: symbolic AI and the connectionist approach.

[*]Corresponding author, Ph.D. Student, E-mail: jdemiguel@us.es
[a] Professor, E-mail: fsancho@us.es

In general, neural-symbolic models utilize a connectionist approach to extract patterns from raw data (such as images), and then implement a symbolic AI engine that operates upon symbolic expressions of these patterns (Evans, Bošnjak, et al., 2021). This approach has achieved spectacular results for example with problems that can be solved by a differentiable maximum satisfiability solver (MAXSAT) (Wang et al., 2019). In light of these works, it is clear that neural-symbolic methods feature important reasoning capabilities. However, among other broader issues (Marcus, 2020), these models need to be trained with labeled data in one way or another, which require immense resources and efforts that limit their scope of applicability. Additionally, this requirement also implies that the method may face substantial challenges in addressing the SGP down the line. In this regard, it is important to note that some neural-symbolic works have addressed the SGP by 'guessing' the labels of the patterns identified by the neural nets, through the imposition a set of constraints. One of such explorations can be found in the paper by Topan, Rolnick and Si (Topan et al., 2021), where the authors use the rules of the Sudoku game as constraints, and claim to have solved the SGP. It would be interesting to review this claim in light of the zero semantical commitment condition or 'Z condition' as articulated by Taddeo and Floridi (Taddeo & Floridi, 2005). In their own words, "The SGP concerns the possibility of specifying *precisely how* an AA [Artificial Agent] can *autonomously* elaborate its own semantics for the symbols that it manipulates and do so from scratch". In any case, it would be desirable perhaps to continue the exploration of the SGP with methods that are more fundamentally generic in their foundation.

In contrast with the neural-symbolic approach, the method proposed in this paper is fully symbolic. It comprises two distinct parts, (i) a sensor where raw data is recursively encoded into symbols, and (ii) a cognitive model based on Formal Concept Analysis (FCA) theory and methodology (Wille, 1982), that generates concept structures from those symbols.

As a first exploration of the method, sensor data will be treated as a multivariate time series of both numerical and symbolic values. This implies that in this first exploration, the objects that can be sensed are trajectory-like. For example, the trajectories of a group of particles immersed in a fluid, or more generally, any one-dimensional geometry that may be obtained for instance, from the spatial contours of physical objects. At the sensor, numerical data is converted into symbols through a minimal (or atomic) feature extraction operation. This process is inspired by Bateson's notion that an idea is in essence nothing but a difference: "a difference that makes a difference" (Bateson, 1999). Similar to the approach presented by Cárdenas-García and Ireland (Cárdenas-García, 2022;

Cárdenas-García & Ireland, 2020), the proposed sensor produces qualitative information by encoding the quantitative aspect of continuous values through a comparator element. In this sense, the most minimal or atomic feature that can be extracted from a stream of numerical data is whether a pair of values within the stream are equal or different. And in being different, which one is larger or smaller than the other. The challenge for the method presented here, is that from just these basic comparison operations, it must be able to build sufficiently rich conceptual representations of the world. This in turn would allow for pattern recognition tasks as in (Cho et al., 2019), and even generative design methods such as (Tofighi et al., 2022). As a backdrop to this approach, it may be discussed that the act of sensing does not understand much of quantities but rather, it understands more about differences. As humans, we are very good at detecting differences, but quite clumsy at measuring magnitudes, and for this precise reason, humans have developed tools that actually measure them for us. Therefore, Bateson's ideas might constitute a good corner stone to build powerful cognitive models. Additionally, because of the atomic nature of the comparisons computed at the sensor, it may be argued that they do not break the Z condition of the SGP discussed earlier. This is a critical distinction because there are quite a number of early works that tackle time series in a very similar way to the method proposed in this paper. In particular, the field of language summarization of time series (Aoki & Kobayashi, 2016; Castillo-Ortega et al., 2011; Kacprzyk et al., 2007), which also engages in the extraction of relatively simple features from data streams, so that they can be expressed semantically in human readable terms. However, there are at least two important differences with the present work: (i) the features extracted in these and other works (Baydogan & Runger, 2014), require prior knowledge of symbolic relationships between variables (for example, a 45º angle as relationship between 'x' and 'y' coordinates, or 'x' being larger than a certain number, etc…), and (ii) the model proposed here generates a hierarchical structure of formal concepts from the features extracted, allowing for multiple conceptualizations of the same object.

As mentioned before, this method creates conceptual structures from atomic features extracted in the sensor. The term 'concept' has many definitions across a wide number of disciplines (Goguen, 2005). In Computer Science, several concept theories have been put forth over the years. In line with the work of Pitts and McCulloch mentioned earlier, the early cyberneticians such as Wiener referred to the notion of 'universals' to capture "what makes a square a square" (Wiener, 1948b). The connectionist avenue gravitated more towards understanding concepts or 'universals' as patterns. Pattern recognition gained enormous traction exhibiting very powerful results, but as discussed earlier, it was limited to the distribution of training data and facing important challenges in terms of explainability and compositionality. However, most approaches to concept learning have been proposed from the standpoint of symbolic AI and feature a strong emphasis on semantics. A good review of these approaches is discussed in the work of Goguen, (Goguen, 2005), which includes the geometrical conceptual spaces of Gärdenfors, the symbolic conceptual spaces of Fauconnier, the information flow of Barwise and Seligman, the formal concept analysis (FCA) of Wille, the lattice of theories of Sowa, and the conceptual integration of Fauconnier and Turner. Although in these theories concepts can be seen as patterns (as in the connectionist approach), they have an inherent symbolic structure that facilitates great compositionality (even the creation of fictions), out-of-distribution learning and explainability (and therefore also, semantics and communication). Thus, the notion of a concept in AI calls for all these features. It is not just a pattern, but a generalization (arguably the most important trait) containing a certain amount of reasoning about the rules and relationships between parts that compose it. For example, when answering the question of what makes a square a square, the answer will invariably carry some reasoning about angles or other geometrical properties. Furthermore, it should be a pattern that can be composed of other patterns, all of which can be explained and communicated in a meaningful way. A way that for example, allows to generalize to other samples yet unseen, similarly to how human concepts work. Because of the very close similarities of the concept structures sought in this work to those of FCA, it has been chosen for the first iteration of the method.

FCA provides with a pattern learning engine, powerful reasoning capabilities and a hierarchical articulation of a concept network. It should come as no surprise to find recent research exploring the potential of this method in the field of explainable AI (Borrego-Díaz & Galán Páez, 2022). However, these structures often result in very large and complex lattices that are very heavy in terms of computation and additionally, they contain such a large number of irrelevant concepts that they become quite inoperable. For this reason, implementations on very detailed data (such as raw sensor data) have remained a constant challenge within the field, with many studies centered on complexity reduction. Some of the early approaches in this regard were led by Radim Bělohlávek exploring constraints based on attribute-dependency formulas (Bělohlávek & Sklenář, 2005a), attribute equivalence (Bělohlávek et al., 2004), hierarchically ordered attributes (Bělohlǎvek et al., 2004) and reduction of fuzzy lattices using hedges (Bělohlávek & Vychodil, 2005). An important line of work in complexity reduction to-date, was the introduction of Granular Computing (GrC) (Yao, 2000; Zadeh, 1979) to the realm of FCA (Bělohlávek & Sklenář, 2005b; Wu et al., 2009). The angle of GrC also allows for a great deal of the flexibility that FCA lacks when dealing with continuous numerical data. Indeed, beyond the issue of computational complexity, FCA is naturally prepared to deal only with discrete symbolic data. When it comes to continuous values, a mapping to symbolic information needs to be provided, or in other words, a set of symbols need to be 'grounded' in the data. The most elementary of these mappings would be simply setting up threshold values that represent specific symbols, for example: {0-25 : A, 25-50 : B, 50-75 : C, 75-100 : D}. Obviously, this method presents numerous limitations, but two in particular. First, the arbitrariness of the thresholds, and secondly the fact that hard limits are rare in natural systems. The GrC approach can help alleviate the first limitation by relying on unsupervised machine learning techniques to automatically generate 'granules' or clusters according to the degree of complexity desired. To overcome the second limitation, apart from variable threshold models (Zhang et al., 2007), Rough Concept Analysis (Kent, 1996; Saquer & Deogun, 1999) and Fuzzy Concept Analysis (Saquer & Deogun, 2001) emerged as important contributions to the field and are widely adopted today. In the end, complexity reduction in formal lattices boils down to none other than heuristics, and therefore the approaches, techniques and methods tend to expand indefinitely with time. According to (Dias & Vieira, 2015), these can be classified into three distinct categories: (i) redundant information removal, which aims to find a minimum lattice in terms of objects and attributes with the same structure as the original, (ii) simplification of the lattice seeking a high-level abstraction that preserves its essential aspects, and (iii) selection of concepts, objects or attributes based on their relevance.

When dealing with more continuous data such as spatio-temporal values, some of the most interesting possibilities may arise in the way attributes are 'selected' according to the modelling strategy, and not just relevance. For example, in the traffic trajectories data modeled in (Almuhisen et al., 2019), constraints are imposed through the aforementioned GrC, but also by creating attributes only from the sequential transitions of the observations rather than looking at all the points of a trajectory as a whole. This heuristic method of mapping or 'grounding' attributes to spatio-temporal data is a constraint strategy based on attribute selection that simplifies the model substantially while successfully extracting useful knowledge. Constraining attribute creation to sequential observations is precisely one of the pillars of the methodology presented here. Despite the proposals above, and also some other more recent ones, to improve the computing efficiency (Mouakher et al., 2021), add flexibility (Min & Kim, 2019) and reduce complexity in concept lattices (Aragón et al., 2021; Hao et al., 2021), the literature throughout the last years is sparse on FCA studies engaging with complex sensor data except for a few examples such as (Boukhetta et al., 2020), where FCA is used to mine sequential patterns in interval-based sequences.

## 2. Method

In what follows, a segment will be considered to be a named tuple of the form $(a_1 = v_1, \ldots, a_k = v_k)$, where each $v_i \in \mathbb{R}$ is the value that the attribute $a_i$ of the segment can take. Non-real values (symbolic, vector, complex, etc.) could be considered, but they will not be addressed in this first approximation and shall be left for future extensions of the model.

A curve/trajectory is a succession of segments that have the same structure (the same set of attributes).

For example, and as will be explained in the next subsection, a segment could be defined from the structure {width, angle}, so a curve would be a succession of segments in which each of them has as attributes (i) the width of the curve in that segment, and (ii) the angle it forms (with respect to a prefixed reference system).

Following the objective of focusing on the successive differences in the curve, three possible comparators will naturally be associated for each of the possible attributes. The comparators used depend on the type of attribute considered; since numerical attributes are considered in this case, all attributes make use of the same type of comparators. Specifically, the comparators $(<, >, =)$ will be used, with the usual semantics when working on real numbers. It should be noted that these comparators are disjoint and complete, i.e., for each pair of possible values in the attributes, one and only one of the comparators is applicable.

Starting from the curve $\alpha := \{s_j\}_{0 \leq j \leq N}$, where the structure of each segment is $\{a_1, \ldots, a_k\}$, the curve feature extraction algorithm consists of the following steps:

**Step 1.** Preprocessing

For each $j < N$ the comparison $s_j$ with $s_{j+1}$ of each attribute is computed, and the interval $[j, j+1]$ is constructed, which, instead of 'k' attributes, shows 'k' comparisons. When engaging with attribute $a_i$ and if $(s_{j+1}.a_i < s_j.a_i)$, the change or transition will be noted as '$a_i$: <' in the interval $[j, j+1]$ ('s.a' denotes the value of attribute 'a' in the segment 's').

The set of intervals with attributes above will be henceforth denoted by $\mathbb{I}\alpha$.

**Step 2.** Compute symbolic differences

Two intervals are contiguous if they are of the form $I_1 = [a, b]$ and $I_2 = [b, c]$. For each pair of contiguous intervals, a new interval is created from their union: $I_1 \cup I_2 = [a, c]$. Then for each attribute of the structure, the transition that occurs from comparisons of the first interval to the second, is computed and associated to the new union interval. In the case of numerical values, nine possible combinations occur, which are noted for convenience as shown in Table 1 (it may be observed that an absorption effect occurs when two comparisons coincide):

Table 1. Symbolic comparison scheme of the nine possible combinations

| $I_1$ | $I_2$ | $I_1 \cup I_2$ |
|---|---|---|
| < | < | < |
| < | = | < = |
| < | > | < > |
| = | < | > < |
| = | = | = |
| = | > | = > |
| > | < | > < |
| > | = | > = |
| > | > | > |

The set of new intervals generated (unions of contiguous intervals of $\mathbb{I}\alpha$) is denoted by $\mathbb{I}^1\alpha$. The set formed by (i) the original intervals and (ii) the potential unions of these intervals computed in this step, is then expressed as $\mathbb{I}^2\alpha \leftarrow \mathbb{I}\alpha \cup \mathbb{I}^1\alpha$.

**Step 3.** Remove redundancies

The information redundancy between contained intervals is checked. An interval $I_1 = [a, b]$ is contained in an interval $I_2 = [c, d]$ if it is verified that $c \leq a \leq b \leq d$. In this case, it is stated that $I_1$ is redundant if for each attribute, the notation of both intervals is the same.

In this step, those intervals that are redundant with respect to another existing interval in the set of intervals shall be removed from the set. Thus, the set $\mathbb{I}^3\alpha \leftarrow \{I \in \mathbb{I}^2\alpha : I \text{ not redundant}\}$ is obtained.

**Step 4.** Recursion

If $\mathbb{I}^3\alpha \neq \mathbb{I}\alpha$, then we make $\mathbb{I}\alpha \leftarrow \mathbb{I}^3\alpha$, and repeat from step 2.

If $\mathbb{I}^3\alpha = \mathbb{I}\alpha$, then the algorithm stops and returns $\mathbb{I}\alpha$ which is used as a formal context for FCA calculation. Through this calculation, both a concept set $\mathbb{C}\alpha$ and a concept lattice $\Gamma\alpha$ are obtained.

In the following paragraph a complete example of application of the algorithm on a specific curve is presented.

### 2.1 Example

In this example a curve is represented in Fig. 1a, with an axis in red color and variable thickness as illustrated by the gray shade around it. A sensor is reading the curve at regular intervals as shown in Fig. 1b. For the sake of simplicity these intervals have been fixed to generate as fewer segments as possible, thus the poor definition of the curved sections. However, in practice, these intervals would allow for a very smooth definition of the curve.

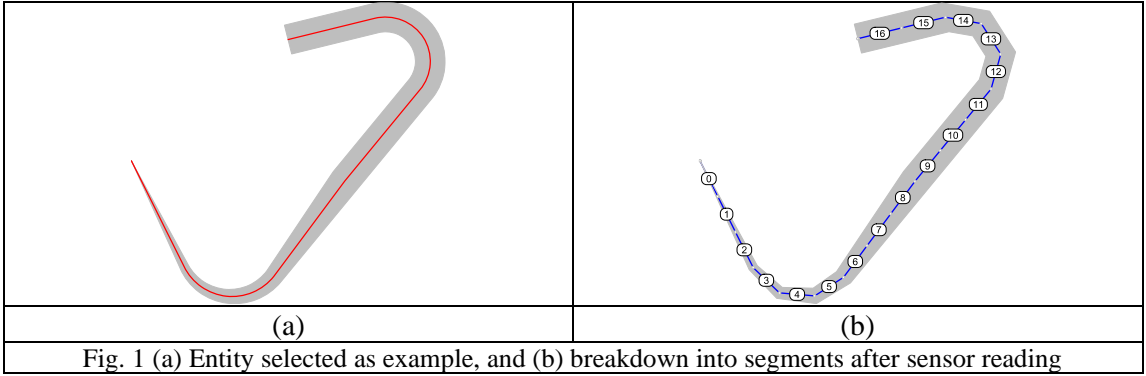

Fig. 1 (a) Entity selected as example, and (b) breakdown into segments after sensor reading

A time-series representation of the curve above, taking into account the angles and the width at each segment is provided in Table 2. It should be noted that the model does not operate with the specific values but rather, it just needs to compare pairs of values and discriminate whether the values are equal, or whether one value is larger or smaller than its peer.

Table 2. Time series representation of the sample entity

| Segment | Width (w) | Angle (a) |
|---|---|---|
| 0 | 0.00 | -65.85 |
| 1 | 0.05 | -65.85 |
| 2 | 0.10 | -65.85 |
| 3 | 0.15 | -46.42 |
| 4 | 0.20 | -7.57 |
| 5 | 0.26 | 31.28 |
| 6 | 0.30 | 50.71 |
| 7 | 0.36 | 50.71 |
| 8 | 0.43 | 50.71 |
| 9 | 0.50 | 50.71 |
| 10 | 0.50 | 50.71 |
| 11 | 0.50 | 50.71 |
| 12 | 0.50 | 74.56 |
| 13 | 0.50 | 122.27 |
| 14 | 0.50 | 169.98 |
| 15 | 0.50 | 193.84 |
| 16 | 0.50 | 193.84 |

**Step 1.** Preprocessing

In the first step, value comparisons of sequential record pairs are run to obtain symbolic properties. The first pair takes the first two segments ($s_0$ and $s_1$) and their values are compared as follows:

$s_0$ {a = -65.85, w = 0.00}
$s_1$ {a = -65.85, w = 0.05}
Compare ($s_1$, $s_0$) = {a: =, w: >}

Applying this operation to all the pairs of contiguous segments results in the following formal context or property table (Table 3):

Table 3. Formal context table resulting from step 1

| Interval | a: = | a: > | w: = | w: > |
|---|---|---|---|---|
| [0,1] | 1 | 0 | 0 | 1 |
| [1,2] | 1 | 0 | 0 | 1 |
| [2,3] | 0 | 1 | 0 | 1 |
| [3,4] | 0 | 1 | 0 | 1 |
| [4,5] | 0 | 1 | 0 | 1 |
| [5,6] | 0 | 1 | 0 | 1 |
| [6,7] | 1 | 0 | 0 | 1 |
| [7,8] | 1 | 0 | 0 | 1 |
| [8,9] | 1 | 0 | 0 | 1 |
| [9,10] | 1 | 0 | 1 | 0 |
| [10,11] | 1 | 0 | 1 | 0 |
| [11,12] | 0 | 1 | 1 | 0 |
| [12,13] | 0 | 1 | 1 | 0 |
| [13,14] | 0 | 1 | 1 | 0 |
| [14,15] | 0 | 1 | 1 | 0 |
| [15,16] | 1 | 0 | 1 | 0 |

Table 4. Resulting concept set after step 1

| Concept | Intent | Extent |
|---|---|---|
| $C_1$ | { } | [0,16] (all) |
| $C_2$ | {a: >} | [2,6], [11,15] |
| $C_3$ | {a: >, w: =} | [11,15] |
| $C_4$ | {a: >, w: >} | [2,6] |
| $C_5$ | {a: =, w: =} | [9,11], [15,16] |
| $C_6$ | {a: =, w: >} | [0,2], [6,9] |
| $C_7$ | {a: =} | [0,2], [6,9], [15,16] |
| $C_8$ | {w: =} | [9,16] |
| $C_9$ | {w: >} | [0,9] |
| $C_{10}$ | {a: =, a: >, w: =, w: > } | - |

The execution of FCA on the previous formal context generates the concept set presented in Table 4. The extent of each concept in the set (except $C_1$ and $C_{10}$) are visualized in Fig. 2.

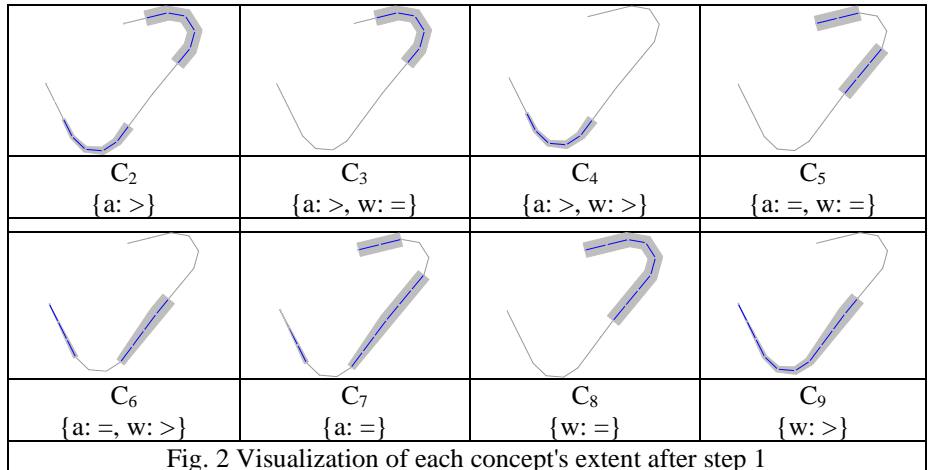

Fig. 2 Visualization of each concept's extent after step 1

**Step 2**. Compute symbolic differences

Next, all contiguous intervals from the previous step are compared according to their symbolic properties. When these properties are the same (i), then there is no transition. Conversely, when there is a difference in properties (ii), the transition is expressed in the form of a new symbolic property. In both cases, a new interval is generated carrying either the preexisting symbolic property or the newly generated one. Below is an example for each case, where the transition from (a: =) to (a: >) is expressed as (a: =>):

(i)  [0,1]: {a: =, w: >}
     [1,2]: {a: =, w: >}
     Compare ([0,1], [1,2]) = [0,2]: {a: =, w: >}

(ii)    [1,2]: {a: =, w: >}
        [2,3]: {a: >, w: >}
        Compare ([1,2], [2,3]) = [1,3]: {a: =>, w: >}

Repeating this process for all the intervals obtained in step 1, results in a new formal context that expands on the previous one presented in Table 3. The new expanded context is presented in Table 5, where the data being repeated has been mostly omitted.

Table 5. Formal context table resulting from step 2

| Interval | w: = | w: > | a: = | a: > | a: => | a: >= | w: >= |
|---|---|---|---|---|---|---|---|
| [0,1] | 0 | 1 | 1 | 0 | 0 | 0 | 0 |
| [1,2] | 0 | 1 | 1 | 0 | 0 | 0 | 0 |
| … | … | … | … | … | … | … | … |
| [14,15] | 1 | 0 | 0 | 1 | 0 | 0 | 0 |
| [15,16] | 1 | 0 | 1 | 0 | 0 | 0 | 0 |
| [0,2] | 0 | 1 | 1 | 0 | 0 | 0 | 0 |
| [1,3] | 0 | 1 | 0 | 0 | 1 | 0 | 0 |
| [2,4] | 0 | 1 | 0 | 1 | 0 | 0 | 0 |
| [3,5] | 0 | 1 | 0 | 1 | 0 | 0 | 0 |
| [4,6] | 0 | 1 | 0 | 1 | 0 | 0 | 0 |
| [5,7] | 0 | 1 | 0 | 0 | 0 | 1 | 0 |
| [6,8] | 0 | 1 | 1 | 0 | 0 | 0 | 0 |
| [7,9] | 0 | 1 | 1 | 0 | 0 | 0 | 0 |
| [8,10] | 0 | 0 | 1 | 0 | 0 | 0 | 1 |
| [9,11] | 1 | 0 | 1 | 0 | 0 | 0 | 0 |
| [10,12] | 1 | 0 | 0 | 0 | 1 | 0 | 0 |
| [11,13] | 1 | 0 | 0 | 1 | 0 | 0 | 0 |
| [12,14] | 1 | 0 | 0 | 1 | 0 | 0 | 0 |
| [13,15] | 1 | 0 | 0 | 1 | 0 | 0 | 0 |
| [14,16] | 1 | 0 | 0 | 0 | 0 | 1 | 0 |

**Step 3.** Remove redundancies

At this point, each interval that is contained in any other interval is checked for redundancy; if the attributes of the shorter interval are the same as the attribute set of the larger, then the former is removed from the interval table. When an interval is removed from the table it is no longer used in the process going forward. For example, the intervals [0,1] and [1,2] are contained in the interval [0,2] and also share the exact same attribute set: {a: =, w: >}. Therefore, both intervals [0,1] and [1,2] are removed from the interval table. However, the interval [15,16] which is contained in [14,16] cannot be removed because their attribute sets are different: {a: =, w: =} ≠ {w: =, a: >=}. After executing this process for all the intervals in Table 5, a new formal context is obtained (Table 6) where redundant intervals have been removed.

Table 6. Formal context after step 3 (removed redundant intervals)

| Interval | w: = | w: > | a: = | a: > | a: => | a: >= | w: >= |
|---|---|---|---|---|---|---|---|
| [15,16] | 1 | 0 | 1 | 0 | 0 | 0 | 0 |
| [0,2] | 0 | 1 | 1 | 0 | 0 | 0 | 0 |
| [1,3] | 0 | 1 | 0 | 0 | 1 | 0 | 0 |
| [2,4] | 0 | 1 | 0 | 1 | 0 | 0 | 0 |

| Interval | | | | | | | |
|---|---|---|---|---|---|---|---|
| [3,5] | 0 | 1 | 0 | 1 | 0 | 0 | 0 |
| [4,6] | 0 | 1 | 0 | 1 | 0 | 0 | 0 |
| [5,7] | 0 | 1 | 0 | 0 | 0 | 1 | 0 |
| [6,8] | 0 | 1 | 1 | 0 | 0 | 0 | 0 |
| [7,9] | 0 | 1 | 1 | 0 | 0 | 0 | 0 |
| [8,10] | 0 | 0 | 1 | 0 | 0 | 0 | 1 |
| [9,11] | 1 | 0 | 1 | 0 | 0 | 0 | 0 |
| [10,12] | 1 | 0 | 0 | 0 | 1 | 0 | 0 |
| [11,13] | 1 | 0 | 0 | 1 | 0 | 0 | 0 |
| [12,14] | 1 | 0 | 0 | 1 | 0 | 0 | 0 |
| [13,15] | 1 | 0 | 0 | 1 | 0 | 0 | 0 |
| [14,16] | 1 | 0 | 0 | 0 | 0 | 1 | 0 |

**Step 4.** Recursion

Steps 2 and 3 are repeated recursively until no new information is generated by the model. This occurs in the present example after three iterations, yielding the final formal context presented in Table 7. From this formal context, 28 concepts are calculated by the FCA algorithm, nine of which have already been obtained in step 1 ($C_1$ to $C_9$). The concept $C_{10}$ is recalculated as it must contain all attributes in the context. In Table 8, the remaining 19 concepts are listed, and their extents are represented in Fig. 3Fig. 5. The resulting concept lattice can be found in Fig. 6.

Table 7. Final formal context obtained

| Interval | w:= | w:> | a:= | a:> | a:=> | a:>= | w:>= | a:=>= | a:>=> | a:=>=> | a:>=>= | a:=>=>= |
|---|---|---|---|---|---|---|---|---|---|---|---|---|
| [15,16] | 1 | 0 | 1 | 0 | 0 | 0 | 0 | 0 | 0 | 0 | 0 | 0 |
| [0,2] | 0 | 1 | 1 | 0 | 0 | 0 | 0 | 0 | 0 | 0 | 0 | 0 |
| [9,11] | 1 | 0 | 1 | 0 | 0 | 0 | 0 | 0 | 0 | 0 | 0 | 0 |
| [0,6] | 0 | 1 | 0 | 0 | 1 | 0 | 0 | 0 | 0 | 0 | 0 | 0 |
| [0,9] | 0 | 1 | 0 | 0 | 0 | 0 | 0 | 1 | 0 | 0 | 0 | 0 |
| [0,11] | 0 | 0 | 0 | 0 | 0 | 0 | 1 | 1 | 0 | 0 | 0 | 0 |
| [0,15] | 0 | 0 | 0 | 0 | 0 | 0 | 1 | 0 | 0 | 1 | 0 | 0 |
| [0,16] | 0 | 0 | 0 | 0 | 0 | 0 | 1 | 0 | 0 | 0 | 0 | 1 |
| [2,6] | 0 | 1 | 0 | 1 | 0 | 0 | 0 | 0 | 0 | 0 | 0 | 0 |
| [2,9] | 0 | 1 | 0 | 0 | 0 | 1 | 0 | 0 | 0 | 0 | 0 | 0 |
| [2,11] | 0 | 0 | 0 | 0 | 0 | 1 | 1 | 0 | 0 | 0 | 0 | 0 |
| [2,15] | 0 | 0 | 0 | 0 | 0 | 0 | 1 | 0 | 1 | 0 | 0 | 0 |
| [2,16] | 0 | 0 | 0 | 0 | 0 | 0 | 1 | 0 | 0 | 0 | 1 | 0 |
| [6,9] | 0 | 1 | 1 | 0 | 0 | 0 | 0 | 0 | 0 | 0 | 0 | 0 |
| [6,11] | 0 | 0 | 1 | 0 | 0 | 0 | 1 | 0 | 0 | 0 | 0 | 0 |
| [6,15] | 0 | 0 | 0 | 0 | 1 | 0 | 1 | 0 | 0 | 0 | 0 | 0 |
| [6,16] | 0 | 0 | 0 | 0 | 0 | 0 | 1 | 1 | 0 | 0 | 0 | 0 |
| [9,15] | 1 | 0 | 0 | 0 | 1 | 0 | 0 | 0 | 0 | 0 | 0 | 0 |
| [9,16] | 1 | 0 | 0 | 0 | 0 | 0 | 0 | 1 | 0 | 0 | 0 | 0 |
| [11,15] | 1 | 0 | 0 | 1 | 0 | 0 | 0 | 0 | 0 | 0 | 0 | 0 |
| [11,16] | 1 | 0 | 0 | 0 | 0 | 1 | 0 | 0 | 0 | 0 | 0 | 0 |

Table 8. Concepts $C_{10}$ - $C_{28}$ (all remaining concepts)

| Concept | Intent | Extent |
|---|---|---|
| $C_{10}$ | {a: =>>=, w: >=, a: =, a: >=>=, a: >, a: =>, a: >=>, a: =>=>=, w: >, w: =, a: =>=>, a: >=} | - |
| $C_{11}$ | {a: >=} | [2,9], [2,11], [11,16] |
| $C_{12}$ | {a: >=, w: =} | [11,16] |
| $C_{13}$ | {a: >=, w: >} | [2,9] |
| $C_{14}$ | {a: >=, w: >=} | [2,11] |
| $C_{15}$ | {a: >=>, w: >=} | [2,15] |
| $C_{16}$ | {a: >=>=, w: >=} | [2,16] |
| $C_{17}$ | {a: =, w: >=} | [6,11] |
| $C_{18}$ | {a: =>} | [0,6], [6,15], [9,15] |
| $C_{19}$ | {a: =>, w: =} | [9,15] |
| $C_{20}$ | {a: =>, w: >} | [0,6] |
| $C_{21}$ | {a: =>, w: >=} | [6,15] |
| $C_{22}$ | {a: =>=} | [0,9], [0,11], [6,16], [9,16] |
| $C_{23}$ | {a: =>=, w: =} | [9,16] |
| $C_{24}$ | {a: =>=, w: >} | [0,9] |
| $C_{25}$ | {a: =>=, w: >=} | [0,11], [6,16] |
| $C_{26}$ | {a: =>=>, w: >=} | [0,15] |
| $C_{27}$ | {a: =>=>=, w: >=} | [0,16] |
| $C_{28}$ | {w: >=} | [0,11], [0,15], [0,16], [2,11], [2,15], [2,16], [6,11], [6,15], [6,16] |

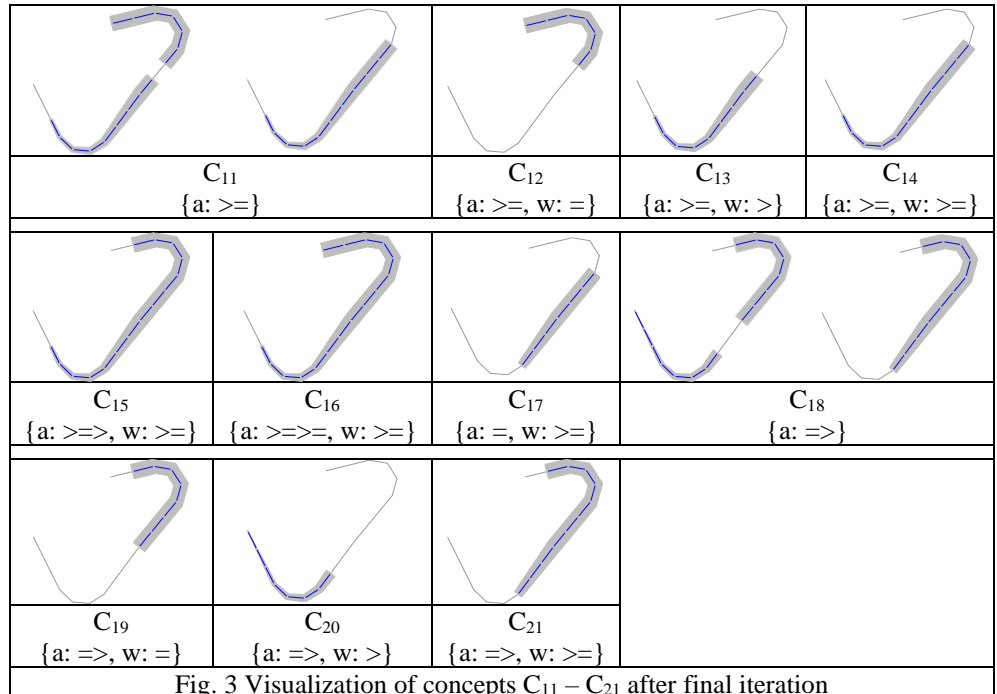

Fig. 3 Visualization of concepts $C_{11}$ – $C_{21}$ after final iteration

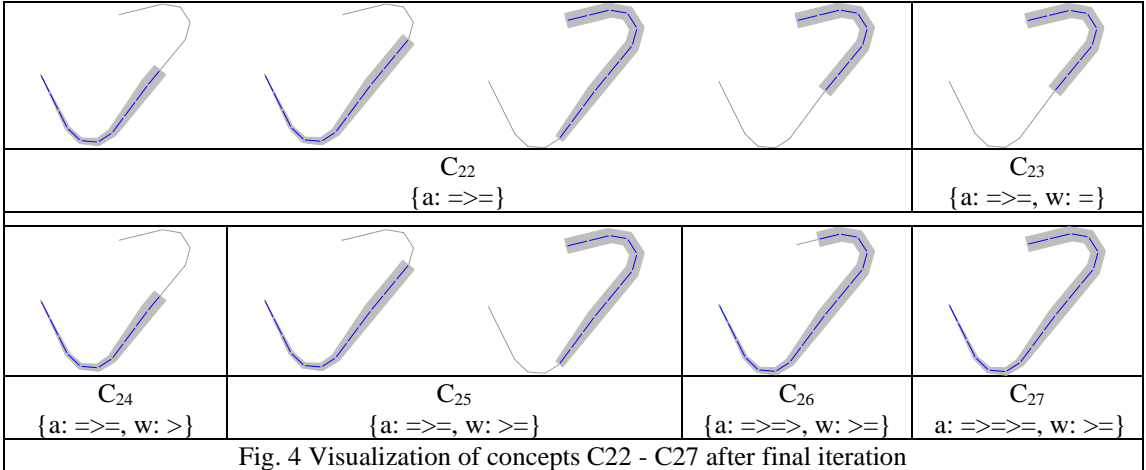

Fig. 4 Visualization of concepts C22 - C27 after final iteration

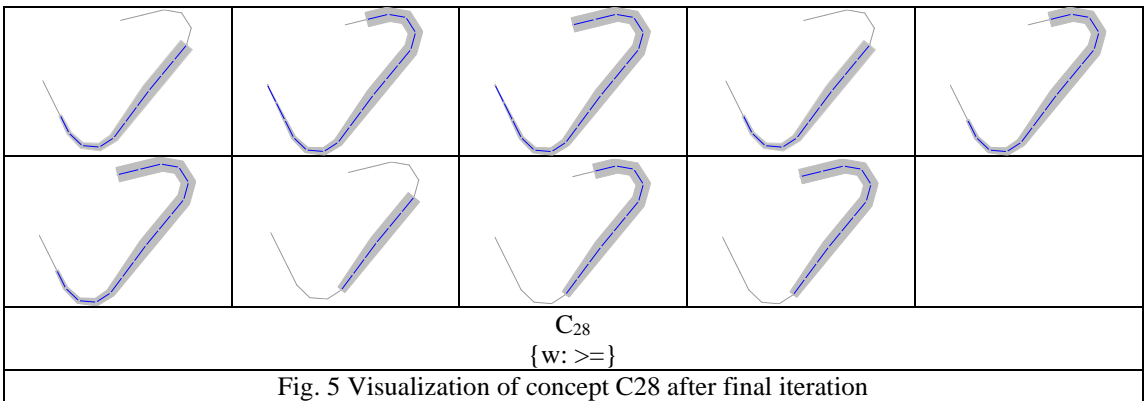

Fig. 5 Visualization of concept C28 after final iteration

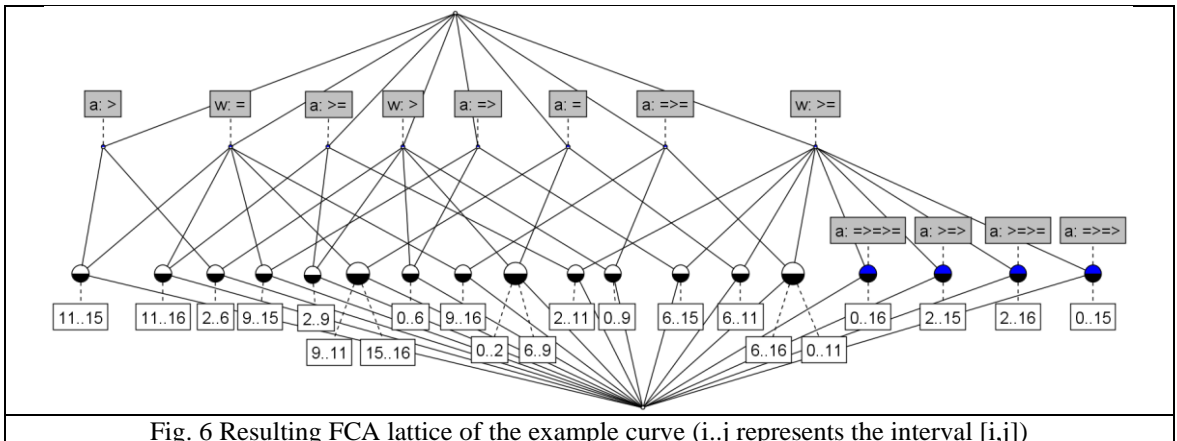

Fig. 6 Resulting FCA lattice of the example curve (i..j represents the interval [i,j])

## 3. Experimentation and results

In this section, the methodology presented above will be assessed in terms of its capability for classification and clustering of a family of curves. Fig. 7 shows the sample-set of curves (or trajectories) that will be considered. All curves feature variable thickness or width, and they flow from left to right. It can be observed that there are four pairs of visually-similar curves (a, b), (c, d), (e, f), and (g, h).

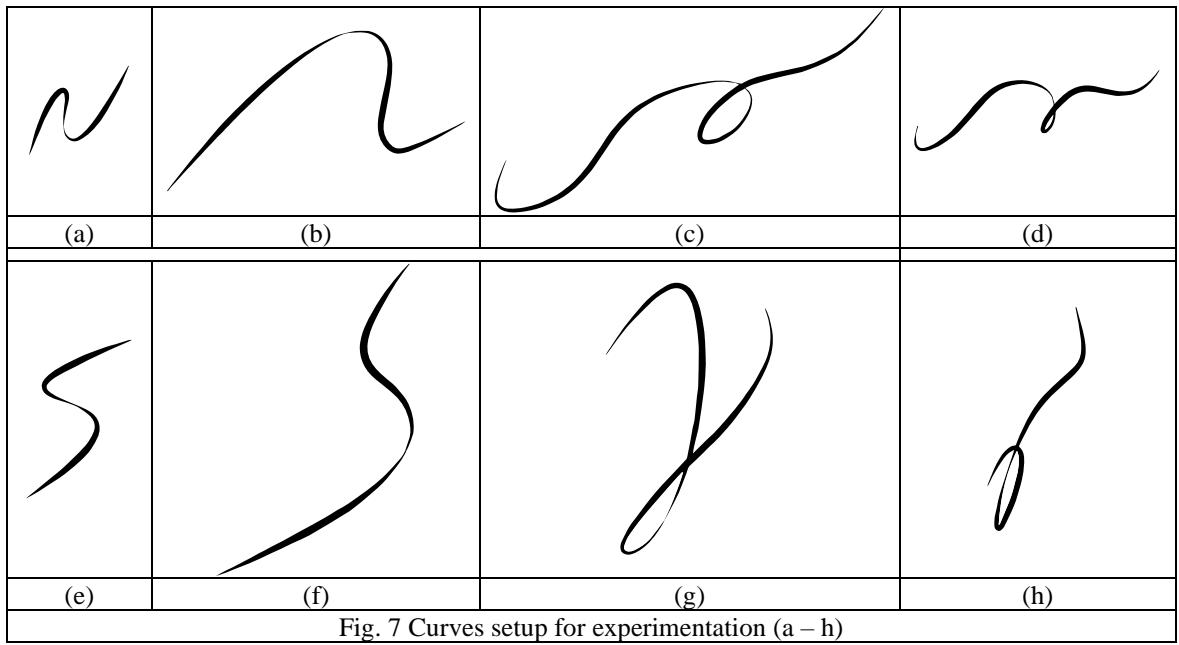

Fig. 7 Curves setup for experimentation (a – h)

The experiments focused on providing answers to three questions: (i) can the model generate concepts that capture common patterns among the curves? (ii) Is there a concept (or set of concepts) that uniquely characterize each of pair of visually-similar curves? And (iii) within each pair, can the model generate a concept that discriminates each individual curve? The results for these questions are presented in the subsections below.

### 3.1 Common patterns across the sample-set

Although this first task is perhaps rather trivial, it allows to further showcase the features of the model. There is of course a great number of concepts that capture common patterns across curves. However, for the sake of simplicity, only those concepts that are common to all the samples are presented (Table 9). A list of concepts has been provided for four different combinations of sensor parameters as follows: {angle}, {angle, width}, {angle, x, y} and {angle, width, x, y}.

Table 9. Concepts (intent) present across all curves in the sample-set

| angle (a) | angle, width (w) | angle, x, y | angle, width, x, y |
|---|---|---|---|
| {a: >} | {a: >, w: <} | {a: >, x: >, y: >} | {a: >, w: <, x: >, y: >} |
| {a: <} | {a: <, w: <} | {a: <, x: >, y: >} | |
| | {a: <, w: >} | | |
| | {a: <, w: > <} | | |

Apart from the concepts provided, this list can be extended to include any subconcepts that can be extracted from them. For instance, if the concept intent {a: >, w: <} is common to all curves, then the subconcepts {a: >} and {w: <} are also common concepts. Just to offer a visual example, the curve sections that form the extent of one of the concepts listed above ({a: <, w: > <}), is shown in Fig. 8.

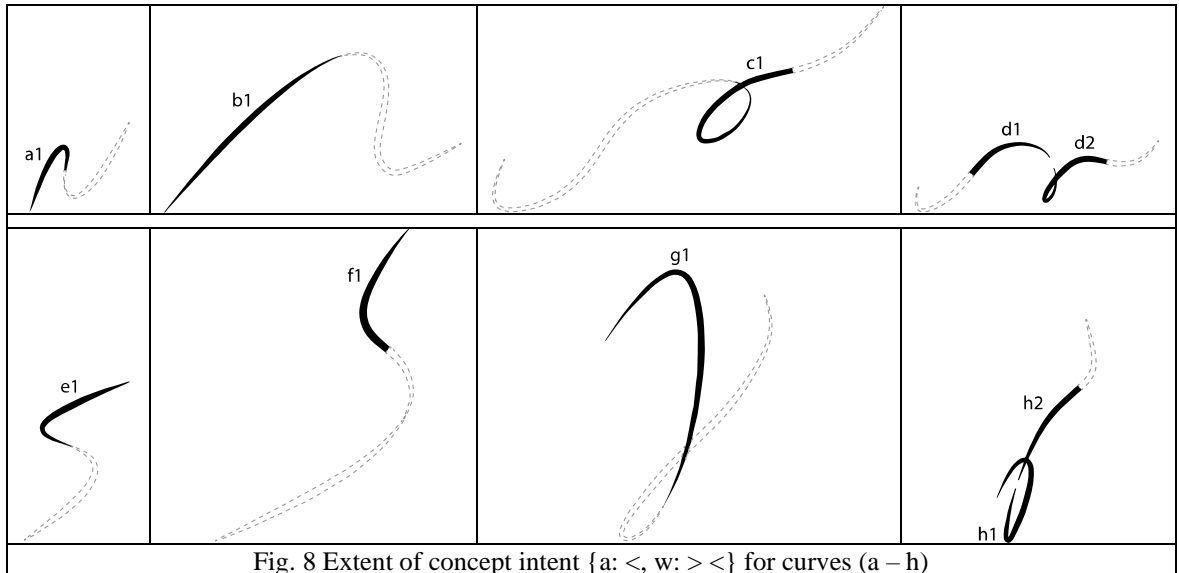

Fig. 8 Extent of concept intent {a: <, w: > <} for curves (a – h)

### 3.2 Characterization of visually-similar curves

For each curve, the model provides the corresponding concept set as per the methodology presented in this paper. A number of different concept sets have been obtained for the following sensor combinations: {angle}, {angle, width}, {angle, x, y}, {angle, x} and {angle, y}. For each combination, the concept set of each curve is compared to that of every other curve in the sample-set. We will denote by $\mathbb{C}\alpha$ the concept set of curve α. If two curves α, β share the exact concept set, then the difference between their corresponding concept sets should be empty: $\mathbb{C}\alpha - \mathbb{C}\beta = \mathbb{C}\beta - \mathbb{C}\alpha = \emptyset$. When one curve's concept set ($\mathbb{C}\alpha$) is a subset of the other's ($\mathbb{C}\beta$), then $\mathbb{C}\alpha - \mathbb{C}\beta = \emptyset$, and $\mathbb{C}\beta - \mathbb{C}\alpha \neq \emptyset$. Finally, if two concept sets are different and none is a subset of the other, then $\mathbb{C}\alpha - \mathbb{C}\beta \neq \emptyset$, and $\mathbb{C}\beta - \mathbb{C}\alpha \neq \emptyset$. The sizes of the differences between concept sets among all curves in the sample-set is presented in matrix form in Tables10-13.

Table 10. Number of differing concepts with parameters: angle

| | | | | angle | | | | |
|---|---|---|---|---|---|---|---|---|
| | a | b | c | d | e | f | g | h |
| a | | 0 | 0 | 0 | 1 | 1 | 0 | 0 |
| b | 0 | | 0 | 0 | 1 | 1 | 0 | 0 |
| c | 2 | 2 | | 0 | 2 | 2 | 2 | 2 |
| d | 2 | 2 | 0 | | 2 | 2 | 2 | 2 |
| e | 1 | 1 | 0 | 0 | | 0 | 1 | 1 |
| f | 1 | 1 | 0 | 0 | 0 | | 1 | 1 |
| g | 0 | 0 | 0 | 0 | 1 | 1 | | 0 |
| h | 0 | 0 | 0 | 0 | 1 | 1 | 0 | |

Table 11. Number of differing concepts with parameters: angle, width

| | | | | angle, width | | | | |
|---|---|---|---|---|---|---|---|---|
| | a | b | c | d | e | f | g | h |
| a | | 4 | 6 | 5 | 9 | 9 | 9 | 9 |
| b | 4 | | 5 | 6 | 9 | 9 | 11 | 9 |
| c | 14 | 13 | | 7 | 12 | 12 | 19 | 15 |
| d | 11 | 12 | 5 | | 12 | 12 | 15 | 10 |
| e | 9 | 9 | 4 | 6 | | 0 | 15 | 15 |
| f | 9 | 9 | 4 | 6 | 0 | | 15 | 15 |
| g | 0 | 2 | 2 | 0 | 6 | 6 | | 0 |
| h | 5 | 5 | 3 | 0 | 11 | 11 | 5 | |

Table 12. Number of differing concepts with parameters: angle, x, y

| | | | | angle, x, y | | | | |
|---|---|---|---|---|---|---|---|---|
| | a | b | c | d | e | f | g | h |
| a | | 0 | 11 | 14 | 48 | 48 | 26 | 26 |
| b | 0 | | 11 | 14 | 48 | 48 | 26 | 26 |
| c | 62 | 62 | | 14 | 95 | 95 | 53 | 53 |
| d | 97 | 97 | 46 | | 128 | 128 | 93 | 93 |
| e | 14 | 14 | 10 | 11 | | 0 | 5 | 5 |
| f | 14 | 14 | 10 | 11 | 0 | | 5 | 5 |
| g | 37 | 37 | 13 | 21 | 50 | 50 | | 0 |
| h | 37 | 37 | 13 | 21 | 50 | 50 | 0 | |

Table 13. Number of differing concepts with parameters: angle, x

| | | | | angle, x | | | | |
|---|---|---|---|---|---|---|---|---|
| | a | b | c | d | e | f | g | h |
| a | | **0** | 2 | 2 | 7 | 7 | 2 | 2 |
| b | **0** | | 2 | 2 | 7 | 7 | 2 | 2 |
| c | 15 | 15 | | **0** | 15 | 15 | 12 | 12 |
| d | 15 | 15 | **0** | | 15 | 15 | 12 | 12 |
| e | 7 | 7 | 2 | 2 | | **0** | 5 | 5 |
| f | 7 | 7 | 2 | 2 | **0** | | 5 | 5 |
| g | 7 | 7 | 4 | 4 | 10 | 10 | | **0** |
| h | 7 | 7 | 4 | 4 | 10 | 10 | **0** | |

As it can be observed in Table 13, the sensor combination {angle, x} produces a unique (and non-subsumable) set of concepts for each visually-similar pair of curves in the sample-set. Thus, the question at stake in this subsection, also renders a positive answer. The concept lattices pertaining to each pair of curves are shown in Figs. 9-12.

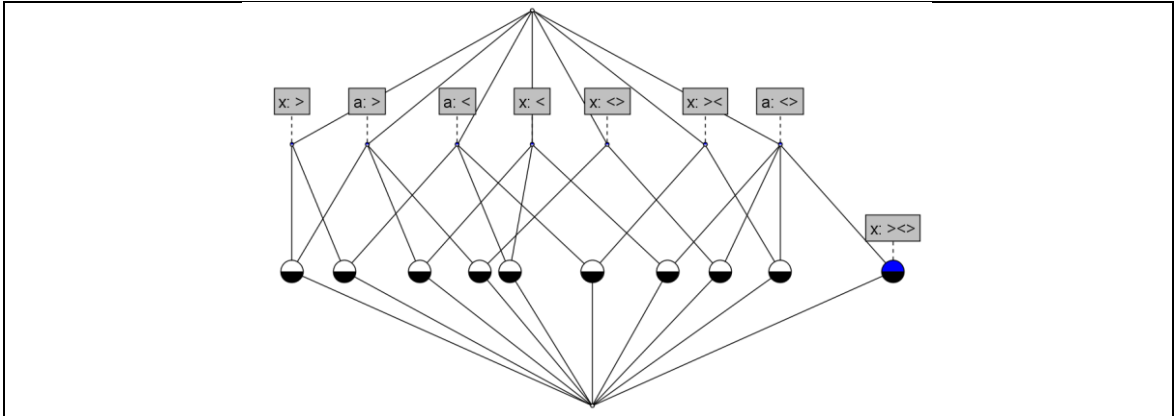

Fig. 9 Concept lattice representation for curves (a) and (b) considering variables {angle, x}

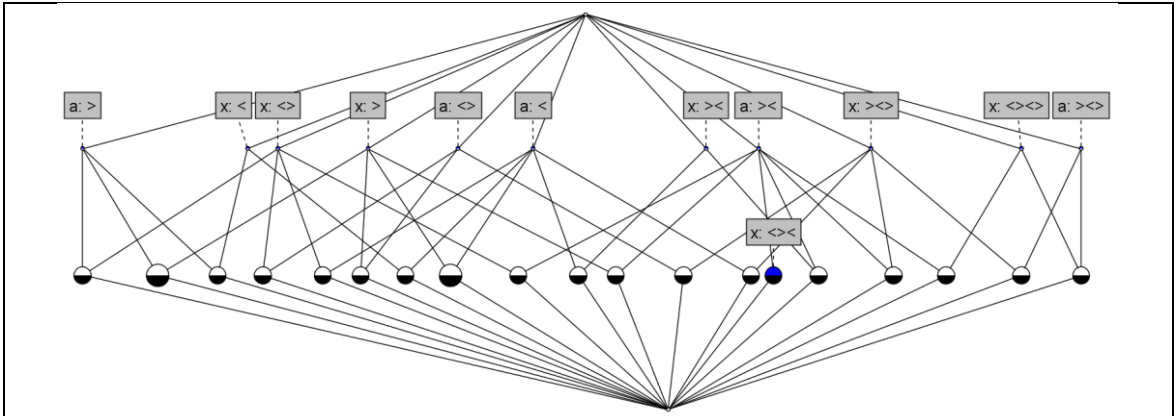

Fig. 10 Concept lattice representation for curves (c) and (d) considering variables {angle, x}

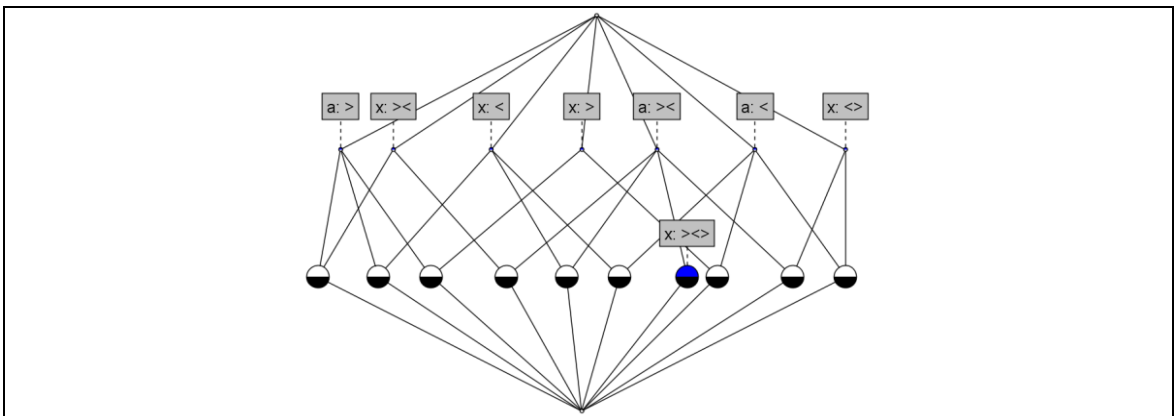

Fig. 11 Concept lattice representation for curves (e) and (f) considering variables {angle, x}

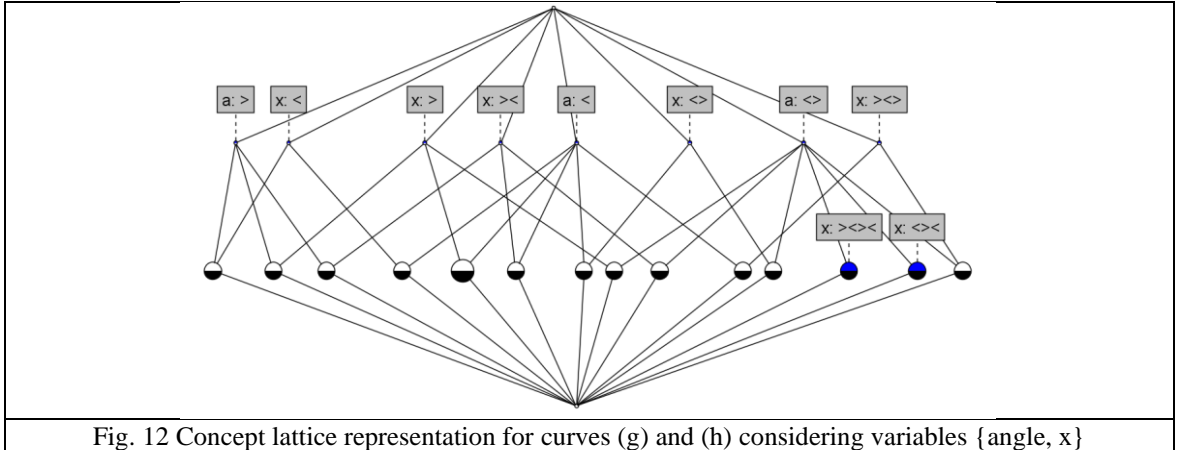
Fig. 12 Concept lattice representation for curves (g) and (h) considering variables {angle, x}

### 3.3 Individual discrimination of curves

For all the curves to be uniquely characterized by a concept set that is also not a subset of that of any other curve, the matrix containing all number differing concepts should not contain any zeros. For the sensor combinations presented in the above subsection, there is no matrix that fulfills this condition; there are conceptual differences among most of the curves but not between all of them. However, after extending the sensor parameters to {angle, width, x, y} a complete discrimination is achieved as presented in Table 14. Thus, in this final case too, the question at stake renders a positive answer.

Table 14. Number of differing concepts with parameters: angle, width, x, y

| | Angle, Width, x, y | | | | | | | |
|---|---|---|---|---|---|---|---|---|
| | a | b | c | d | e | f | g | h |
| a |   | 85 | 79 | 74 | 132 | 135 | 104 | 102 |
| b | 94 |   | 95 | 93 | 136 | 141 | 110 | 113 |
| c | 194 | 201 |   | 65 | 231 | 238 | 189 | 196 |
| d | 250 | 260 | 126 |   | 295 | 302 | 249 | 258 |
| e | 61 | 56 | 45 | 48 |   | 25 | 56 | 51 |
| f | 58 | 55 | 46 | 49 | 19 |   | 56 | 49 |
| g | 94 | 91 | 64 | 63 | 117 | 123 |   | 51 |
| h | 125 | 127 | 104 | 105 | 145 | 149 | 84 |   |

## 4. Discussion

Although the method presented is quite simple in its preliminary form, the three questions addressed in the experimentation section have been answered successfully. Of course, the tests have been performed on a specific sample-set and therefore, these results are only a proof of concept and cannot be generalized. Further experimentation should be carried out with larger datasets like the 'omniglot' (Lake et al., 2015), where proper benchmarking can be executed.

In the first experiment, a number of common concepts have been found across all curves as presented in Table 9. As a machine learning model, pattern recognition abilities are of obvious importance. And owing to the symbolic nature of the model, it can be clearly observed that all those concepts are easily expressed in natural language. A visual example of one such concept has been provided in Fig. 8 (intent: {a: <, w: > <}). Since this concept relies only on width and angle variations, it is rotation and scale invariant. The curve sections (extent) verifying this concept are those presenting a decreasing angle while their thickness first increases and then, decreases. Visually, this seems evident for sections $a_1$, $b_1$, $e_1$, $g_1$ and $h_1$, but less so for sections $c_1$, $d_1$, $d_2$, $f_1$ and $h_2$, where the thickness does not appear to increase or decrease as much (of course, numerically it does). It should also be noted that a visual analogy is probably not the best point of comparison with the model, but more likely, a sort of tactile swiping along the curves could provide a better analogy with human sensing. Finally, the sections $h_1$, $c_1$ and $d_2$ are self-intersecting, which is very visually apparent, but again, that is not really how the model operates. Regardless, a tactile sensing model could still detect these intersections if perhaps, it was provided with sufficient memory. However, this initial version has no substantial memory as of yet, and thus, these features are not detected.

The second question attempted to find unique conceptual definitions for each pair of similar curves. These have been found for the sensor combination (angle, x) as presented above. The first thing to note is that since 'x' is involved, these definitions are not rotation-invariant. Clearly, the angle alone would not be able to discriminate the curves 'e' and 'f' from 'c' and 'd', as the angle patterns of the former are a subset of the latter. Then, thickness as a second discriminator does not follow a similar trend within each pair, thus, not providing the differentiation required. In the method presented, input data is strictly of numeric type. However, further development may consider incorporating symbolic input data as well. This would allow for expressions such as a transition from no sensing to sensing and vice versa. In turn, expressions of this kind may provide further differentiation; especially in some cases like the one mentioned above, where entire curves ('e' and 'f') are a conceptual subset of others (both 'c' and 'd'). Moving on, the sensor combination {angle, x, y} is able to characterize all pairs except 'c' and 'd'. Interestingly, these two curves feature an important conceptual difference: 'd' plunges down after the loop (a decrease in y), while in 'c' the curve is always moving upwards after the loop. Then, the combination {angle, width} is only able to produce a unique definition for the pair of curves 'e' and 'f'.

Another important observation to be made is that the process of finding specific sensor combinations that produce unique characterizations for every pair of curves, has been executed manually. This search can be thought of as a training process and as such, it should be formalized and developed in future work. In the samples shown here, there are only two training samples per category. Also, a single sensor combination is used for an entire curve. An immediate takeaway advantage of the method is that it can train with as few samples as desired, and conceptual knowledge is always generated even without training. On the flip side, large datasets might pose a challenge. However, it would be possible to dynamically adjust the combination of sensor parameters for different sections of the curves as a training parameter. This, perhaps, could offer a viable approach to this challenge.

The last experiment attempted to distinguish all the curves from each other. This was achieved using a sensor combination that includes all the parameters {angle, width, x, y}. Of course, although the experiment was successful with the eight curves in the sample-set, there is a severe limitation on the differentiating capacity of the model in its current form. Given two S-shaped curves for example, one with more accentuated bulges than the other, then it is very likely that the model will not be able to discriminate between the two. The same applies for a very long bulge versus a short one, etc. This difficulty in assessing magnitudes, generating concepts that quantify change in sensor variables, is a very important challenge of this model as illustrated in Fig. 13. However, there are at least three different avenues that may be explored going forward to address it.

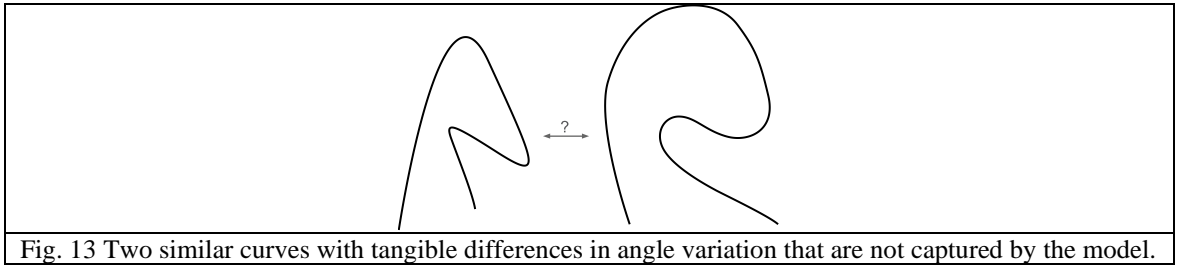
Fig. 13 Two similar curves with tangible differences in angle variation that are not captured by the model.

The first strategy would entail performing arithmetic operations on the values coming through the sensor to gain valuable data. Then, the atomic comparisons of the method can be run on them to extract symbolic properties. This can be done in many different ways. For example, any form of aggregated values (such as number of segments, total length or average thickness, etc.) can be extracted from intervals of the extent of a concept. Subsequently, a concept holding two separate intervals within its extent can be unfolded into two separate concepts. This is the case if for instance, those intervals are of different total length or any aggregate value chosen. Conversely, these values can become part of the standard comparisons of the method rendering concept intents like {a: <, l: =}. Where 'a' is the angle parameter and 'l' is the length of the intervals being compared. Another possibility is to perform these arithmetic operations at the level of the sensor values themselves. For example, the difference in angle between two consecutive segments can be compared to that of the previous two segments. This would produce a symbolic token relating to the magnitude of change in angle. Other operations could include comparing the difference in value of two segments to the actual value of each one. In general, any operation that allows to extract magnitude information in the form of ratios or proportions would fall within the spirit of the model.

A second avenue of development involves creating comparison spaces between objects and drawing new sensory data from their relative differences. For example, two vehicle trajectories may be compared by tracking the distance between the two vehicles over time. Then that distance can be used or computed as yet another sensor parameter. In the case of static curves, these can be placed next to each other in different relative positions (in a similar way to how children play around with toy-shapes observing the effects of different spatial arrangements). Similarly, a distance parameter flowing along them can be integrated into the sensor. In turn, this strategy produces concepts that directly express differences between initially similar curves. Furthermore, these differences can be made extensive not only to pairs of curves but to any finite number of them.

Finally, a third approach may be explored through the implementation of a memory to store specific values. These values in memory could then be compared retrospectively beyond the value pair comparisons described in the present method. This last option is perhaps a bit far removed from the principles of the model. As humans, it is obvious that we do have a better grasp of proportions and ratios rather than specific magnitudes (measuring tools are needed for a reason). Therefore, approaches prioritizing relative magnitudes over absolute ones are perhaps more in-line with the work presented in this paper.

Another limitation of the model (implicit in the experimentation) is that nested concepts obscure their containing ones. For example, a large S-shape that is composed of smaller S-shapes is not recognized by the model (Fig. 14). Such problems could be addressed by more 'embodied' approaches, e.g.: a sensor system with a main arm and a sub-arm (or hand), whereby the main arm moves at a wider scale and the hand captures smaller nuances (including noise).

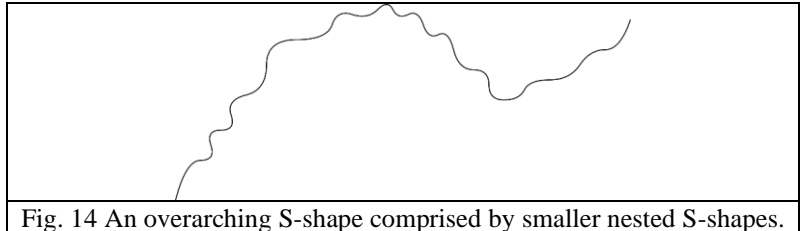
Fig. 14 An overarching S-shape comprised by smaller nested S-shapes.

From the experimentation, it can also be inferred that the model is essentially a generative one. Because it is radically composable, new concepts or features that have never been perceived by the sensor can be formed from combinations of those belonging to past experience. Furthermore, since there is a fully tractable rout from the sensor data to the concepts created by the model, these 'unseen' new features can be propagated or decoded back to the sensor level. Such a decoder would allow to output the corresponding curves for these unseen concepts. Of course, because every concept in the model presents a high degree of generalization, there are infinite potential outputs for each new concept. Thus, further work would be required to establish a heuristic framework responsible for determining the final output (extent) for a given concept. Additionally, the fact that all concepts are easily expressed in natural language implies that (i) sensor data of both static (nouns) and dynamic (verbs) nature from the world is being encoded into human-readable language, and (ii) new objects may be imagined and output to the world by combining existing concepts into new ones through natural language queries. In this sense, it would be interesting to engage in future work that takes up this model as the basis for the development of a natural language engine for sensorial data. Given the recent advances of large language models (LLM), this ambitious avenue could open up enormous possibilities.

Finally, apart from all the pointers to further research already mentioned, there are more important areas of work. The most fundamental one perhaps is that currently, the model presented only processes data in the form of time series or trajectories. The extension to incorporate bidimensional data such as images (or even multidimensional), is definitely not trivial and will require a great deal of effort. Other directions for exploration may include integrating the approach laid out in this paper with more established methods in neural-symbolic research that tackle the same challenges (Evans, Bošnjak, et al., 2021; Evans, Hernández-Orallo, et al., 2021). Or, on a different note, future work could be also carried out to attempt a physical implementation of the model presented here, which in turn could offer a very promising edge with regards to the symbol grounding problem.

## 5. Conclusions

This paper has presented an algorithmic machine learning model to generate human-relatable concepts from spatial sensor data. Its mechanics are based on Bateson's notion that an idea (concept) is in essence, nothing but a difference. In this spirit, the model extracts basic features from the data based on atomic value comparisons (=, >, <), and then, generates more complex concepts by recursively computing differences of previously detected differences. The model has been formulated in formal terms, and is of general applicability. However, although the input data has been laid out as a time series format, the method is intended for spatial sensor data of everyday objects, rather than time series data of more abstract (and complex) nature. This is an important distinction because there is already plenty of research in the area of language summarization of time series. In this sense, the present work differs mainly in that it pursues different objectives; priorities to generalization, composability, flexibility and simplicity are given over accuracy in pattern discrimination.

As laid out in the methodology, the model is inherently built for generalization, out-of-distribution learning and high composability. Moreover, it is provided with generative and reasoning capabilities (powered by FCA). Additionally, results show that it can generate 'fairly' rich representations of the data. They also indicate, that these representations are capable of both discriminating and assimilating objects on demand, by elaborating multiple representations of a same object with extreme ease. Optionally (not necessarily), training processes can also be implemented to further fine-tune these operations.

In a nutshell, this set of traits are the main aspiration fueling the recent explosion in neural-symbolic developments. Neural-symbolic research is overcoming the limitations of highly composable symbolic models with great generalization capabilities but unable to produce rich representations from complex raw data. Connectionist approaches instead, had thoroughly proved this last ability after the rise of 'deep learning'. Thus, the combination of neural and symbolic approaches is rightfully spawning machine learning models that generalize better, achieve out-of-distribution learning to a greater extent, are more composable than their initial deep learning counterparts and yet, are able to generate extremely rich representations. The neural engine in these models however, still requires intense training for the most part, and labeling of data samples continues to pose important challenges for the industry. In contrast, in the method presented in this paper, training is just optional – not strictly required; complete conceptual representations are generated regardless of the number of samples sensed. Furthermore, each of these representations can be easily expressed in natural language without any need for labeling of features. And also, every concept generated can be traced back down to its originating atomic comparisons, adding full explainability to the model in natural language as well. Concepts generated with this method, as discussed earlier, can be of static or dynamic nature; capturing both object and movement-related concepts (nouns and verbs). In light of the explainability aspect just mentioned, this implies that potentially, a basic language could be built bottom-up from primitive tokens such as sensor parameter identifiers and the comparators (=, >, <) used in this work. In fact, one of the most interesting facets of this model is that it creates full semantic structures of complex concepts without any external intervention or label assignment other than those primitive tokens. And this holds true even if the model senses a new object only once and for the first time. Therefore, it may be argued that Bateson's principle can open up a very interesting avenue of study in relation to the symbol grounding problem discussed in this paper. For all these reasons, it may be suggested that the methodology presented here could add important contributions to the field of machine learning. Despite nonetheless, of the obvious difference in the 'richness' of representations achieved in this first and early implementation, when compared to current neural-symbolic works. Nevertheless, although limited, the experimentation section does show promising results. It is acknowledged notwithstanding, that this performance should be formally assessed as soon as possible, by means of proper benchmarking against a well-known dataset. In any case, this gap may perhaps be reduced significantly in the future, by joint efforts of the scientific community.

As a final remark, perhaps Bateson's approach to the generation of concepts from sensor data has not drawn all the attention it deserved. If this methodology can prove to yield sufficiently rich representations, then all the advantages of symbolic models could be harnessed regardless of the support of neural networks in the process of sensing. To the question 'why was this paper not written 20 years ago?' no answer has been found by the authors.